\documentclass{article}
\usepackage[dblblindworkshop, final]{neurips_2026}
\workshoptitle{TAE (Trust-AI-Eval): Can We Trust AI Evaluation?}
\usepackage{graphicx,booktabs,amsmath,amssymb,xcolor,url}
\usepackage{tikz}
\usepackage{placeins}
\usepackage{float}

\title{Are We Measuring Anticipation?\\
Auditing Privileged Information in Procedural Video Evaluation}
\author{
Mahsa Mohammadi\\
University of Exeter\\
Exeter, United Kingdom\\
\texttt{mm1510@exeter.ac.uk}
\And
Sareh Rowlands\\
University of Exeter\\
Exeter, United Kingdom\\
\texttt{S.Rowlands@exeter.ac.uk}
}

\begin{document}
\maketitle

\begin{abstract}
Benchmark scores license claims about the capabilities being evaluated. We audit
the inference licensed by an evaluation protocol, rather than the predictive model
alone. Using procedural action anticipation as a controlled case study, we study a
broader evaluation-validity failure mode: a protocol can remain temporally causal
and free of classical target leakage while still supplying a privileged
intermediate representation. The failure is not merely optimistic accuracy, but a
mismatch between the capability claimed and the construct actually measured.

On Breakfast, matched recognizer-history and GT-history regimes score $30.3\%$
and $62.3\%$
($\Delta_{\mathrm{PH}}(R_{\mathrm{BA}})=+32.0$,
95\% Student-$t$ interval $[+24.1,+39.8]$). A fixed-checkpoint
$2\times2$ intervention isolates a $+14.6$-point test-time oracle contrast; a
matched no-video provenance probe yields a $+26.8$-point gap; and a
boundary-independent query/history control retains a $+19.4$-point gap. We refer to this discrepancy as a \emph{privileged-information gap}, defined
relative to a specified non-oracle recovery pipeline, and propose a reusable
four-condition audit. Across five evaluation settings the gap is heterogeneous;
a standard 50\,Salads long-term-anticipation re-implementation provides cautious
support beyond the audit-specific next-action construction.
\end{abstract}

\section{Introduction}
A benchmark score is scientifically useful only if the evaluation protocol
licenses the capability claim attached to it. This is a measurement and
construct-validity problem
\cite{cronbach1955construct,messick1995validity,raji2021everything,salaudeen2025measurement}.
Temporal validity is necessary but not sufficient: a protocol may exclude future
frames and the target label while still supplying an oracle representation
unavailable to the claimed system.

We study this evaluation-validity failure mode using procedural action anticipation
as a controlled case study. The intended construct is future-action prediction
from non-oracle observations available by prediction time, yet some regimes
provide the observed past as a ground-truth symbolic transcript. Prior work already
compares clean/GT and recognized histories
\cite{farha2019uncertainty,farha2020cycle,benavent2025aag}; our novelty is to treat
history provenance as an evaluation variable and audit the inference supported by
the protocol, rather than the predictive model alone.

On Breakfast, a boundary-aware recognizer-history model reaches $30.3\%$ Top-1,
whereas the matched GT-history regime reaches $62.3\%$, a paired
$+32.0$-point gap. Because these two conditions are trained separately, this is a
protocol-level information-regime gap rather than a pure test-time effect. A
fixed-checkpoint $2\times2$ intervention therefore evaluates the same
recognizer-trained model with recognized versus GT history and isolates a
$+14.6$-point oracle-substitution contrast. A matched no-video probe further
separates context predictability from provenance: recovered context reaches
$32.5\%$ whereas oracle context reaches $59.2\%$. Finally, a fixed-schedule
query control removes annotated boundaries from both query timing and recovered
history construction and still yields a $+19.4$-point gap.

We refer to the score discrepancy induced by substituting a recoverable
intermediate variable with an oracle counterpart as a
\emph{privileged-information gap}. For action history, the operational quantity
$\Delta_{\mathrm{PH}}(R)$ is explicitly relative to recovery pipeline $R$ and is
not an intrinsic scalar property of a dataset. We additionally re-implement the standard
50\,Salads observe-$0.2$/predict-$\beta$ long-term anticipation protocol to check
that the direction is not unique to our diagnostic next-action construction.

Our contributions are:
\begin{itemize}\itemsep1pt
\item we formalize privileged intermediate information as an
\emph{evaluation-validity} problem and define the deployment-relative,
estimator-relative audit quantity $\Delta_{\mathrm{PH}}(R)$;
\item we introduce a reusable four-condition audit that separates primary visual
evidence, recovered symbolic context, oracle symbolic context, and context-only
predictability;
\item we provide an identification hierarchy---matched regimes, fixed-checkpoint
substitution, no-video provenance, and boundary-independent query/history
recovery---that targets distinct rival explanations for the score gap; and
\item we evaluate scope, recovery sensitivity, uncertainty, and standard-protocol
external validity, then translate the findings into a concrete reporting checklist
for benchmark interpretation.
\end{itemize}

\section{Related Work}
\paragraph{Evaluation validity.}
Construct-validity work asks whether a score supports its intended interpretation
\cite{cronbach1955construct,messick1995validity,borsboom2004concept}; recent ML
evaluation work makes the same point for benchmark claims, estimands, and
reliability \cite{raji2021everything,liao2021learning,binette2024estimands,
bouthillier2021variance}. We audit a complementary failure mode: an
annotation-supplied intermediate representation can change the construct for which
the score is evidence.

\paragraph{Anticipation and observed-past representations.}
Anticipation methods predict short- and long-horizon futures from observed video
\cite{furnari2019expect,girdhar2021avt,ke2019time,farha2018when,gong2022futr},
while prior work uses clean/GT, recognized, visual, and language-derived histories
\cite{farha2019uncertainty,farha2020cycle,manousaki2023vlmah,zhao2024antgpt,
benavent2025aag}. We do not claim ``GT $>$ recognized'' as new, nor that prior
work necessarily misinterpreted oracle scores. Our contribution is a formal audit
and identification framework that manipulates provenance as an evaluation variable
and pairs it with fixed-checkpoint, no-video, boundary, recovery-quality, and
uncertainty controls.

\section{Audit Framework}
\paragraph{What inference is being audited?}
The intended construct is anticipation from non-oracle observations; the
oracle-conditioned regime evaluates anticipation given privileged procedural
information. $\Delta_{\mathrm{PH}}(R)$ measures the score discrepancy relative to
recovery pipeline $R$. The audit therefore targets the capability inference
licensed by the protocol, not model accuracy in isolation.

\paragraph{When is an intermediate variable privileged?}
Privilege is relative to the capability claim. We call an intermediate variable
privileged when (i) it is not directly available from ordinary system inputs at
prediction time, (ii) evaluation obtains it from annotation or another oracle
source unavailable under that claim, and (iii) a non-oracle system would have to
recover an estimate from observations. Importantly, we do not argue that
ground-truth histories are inherently invalid evaluation inputs. If the intended
deployment genuinely provides an authoritative action history, evaluating with
ground-truth history is appropriate; if the deployed system must infer that history
from observations, the same evaluation measures performance conditional on
privileged information.

\begin{figure*}[t]
\centering
\begin{tikzpicture}
\node[inner sep=0] (auditfig) {\includegraphics[width=\textwidth]{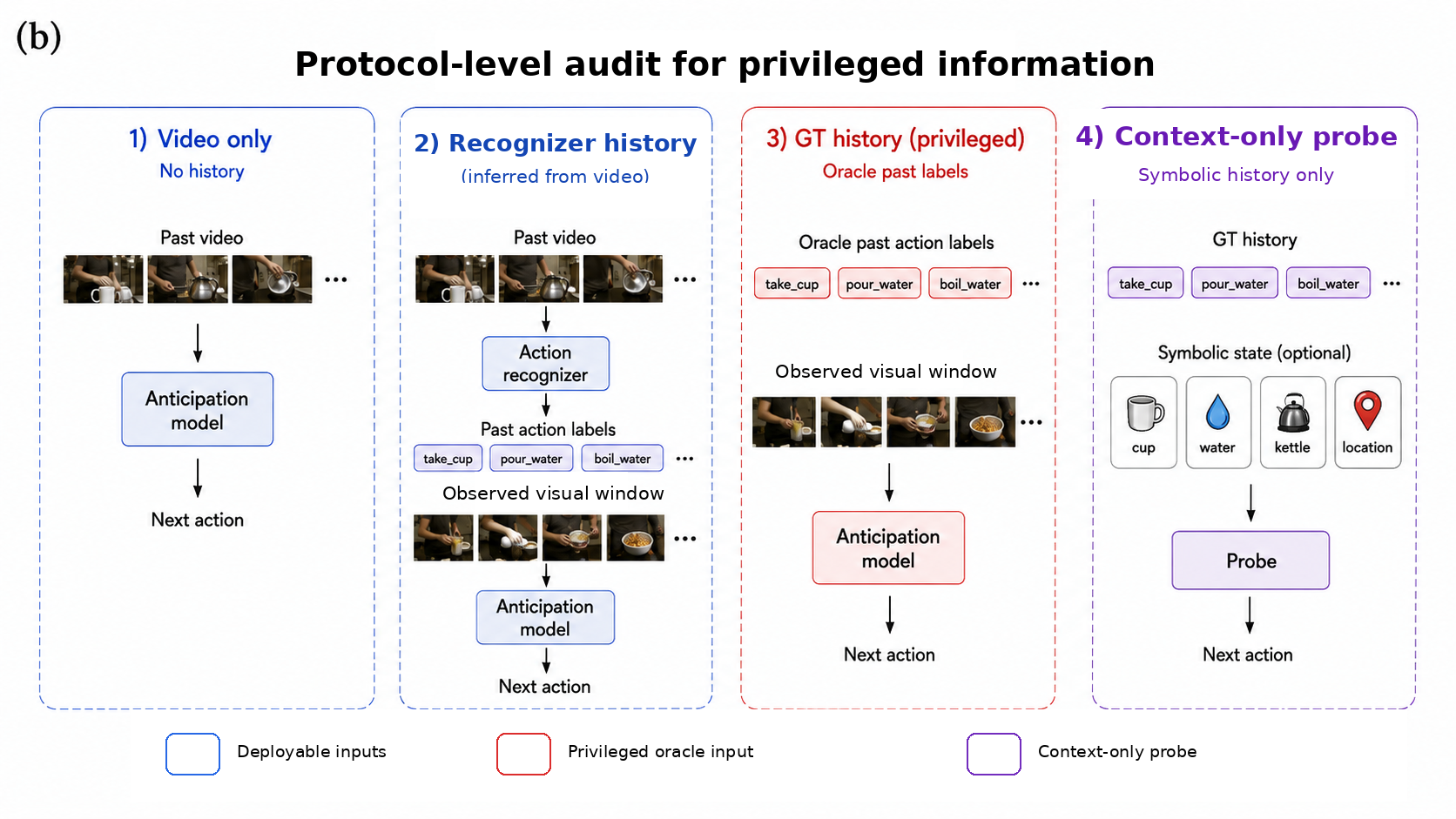}};
% Remove only the obsolete panel marker "(b)".
\fill[white] ([xshift=-0.03cm,yshift=0.03cm]auditfig.north west)
             rectangle ([xshift=0.78cm,yshift=-0.63cm]auditfig.north west);
% Preserve the original legend geometry. Hide only the old "Deployable inputs"
% wording and replace it with the locked terminology.
\fill[white] ([xshift=2.15cm,yshift=0.28cm]auditfig.south west)
             rectangle ([xshift=4.65cm,yshift=1.02cm]auditfig.south west);
\node[anchor=west,inner sep=0pt] at
      ([xshift=2.38cm,yshift=0.64cm]auditfig.south west)
      {\footnotesize Non-oracle inputs};
\end{tikzpicture}
\caption{\textbf{Four-condition audit.} Video only measures the primary modality;
recognizer history adds recoverable symbolic context; GT history supplies the same
semantic variable from an oracle source; and the context-only probe removes the
observation-window visual input. Blue denotes non-oracle inputs; red denotes the
privileged oracle input.}
\label{fig:audit}
\end{figure*}

\paragraph{Estimator-relative audit quantity.}
For split $s$ and recovery pipeline $R$,
\begin{equation}
\Delta_{\mathrm{PH}}^{(s)}(R)=
\mathrm{Acc}^{(s)}(\text{GT-history regime})-
\mathrm{Acc}^{(s)}(\text{non-oracle regime }R).
\end{equation}
The main value is the mean paired gap over official splits. Positive values mean
that the oracle-conditioned score exceeds the matched non-oracle score; negative
values indicate the reverse. Because GT-history and recognizer-history models are
trained under their corresponding information regimes, the matched gap is a
protocol-level estimand, not a pure causal test-time effect.

\paragraph{Task, recovery pipelines, and controls.}
The primary audit uses Breakfast; 50\,Salads, GTEA, Assembly101, and EPIC-100
provide additional evaluation settings. The cross-setting 50\,Salads row is a
boundary-aligned next-action audit setting and is distinct from the separate
standard long-term-anticipation re-implementation reported below. GTEA,
Assembly101, and EPIC-100 use diagnostic next-action protocols constructed for
this audit, not their official anticipation benchmarks. Visual inputs are observed-prefix I3D/TSM/TSN features;
future frames and the target label are never inputs. The boundary-aware recovery
pipeline labels completed observed segments. The stricter boundary-free history
recognizer uses 16-feature-frame windows with stride 8 feature frames and collapses consecutive duplicate
predictions. Main diagnostic query times remain annotation-aligned.

A separate Breakfast control issues queries on the extracted feature sequence,
starting at feature-frame 64 after a 64-feature-frame warm-up and then every
60 feature frames. The target is the first action starting after the query, and
recognizer history is constructed from the fixed-window recognizer. No annotated
segment boundary is used as a model input, to schedule the query, or to construct
recognizer history; annotations are used only to define evaluation targets.
For this control, the recovered symbolic history is truncated/padded to the eight
most recent action labels ($H=8$).

\paragraph{Reference implementation.}
The reference anticipator projects 2048-d I3D/TSM visual features to $d=256$ and
encodes the observed window with a 2-layer transformer encoder (4 heads, GELU,
dropout $0.1$) followed by mean pooling over valid frames. Past actions are embedded
with PAD/SIL tokens and encoded with a GRU. When used, the symbolic state is a
deterministic binary summary of the observed prefix (held-object identity,
processed/contained indicators, and appliance state) and never uses future
annotations. Visual, history, and state representations are concatenated and
passed through a linear+LayerNorm+GELU fusion layer with action, verb, and noun
prediction heads.

Training uses AdamW with weight decay $10^{-4}$, learning rate
$3\times10^{-4}$, batch size 128, and 8 epochs. The action cross-entropy is
augmented with verb/noun losses weighted by $0.3$ each. The main observation
window contains the last 32 feature frames before the query. The boundary-aware history
recognizer is a 3-layer MLP over mean segment features; the stricter
boundary-free recognizer labels 16-feature-frame windows with stride 8 feature frames and collapses
consecutive duplicate predictions. All matched oracle/non-oracle regimes use the
same architecture, optimization, splits, targets, visual construction, and
training budget; the supplement gives the remaining implementation details.

\paragraph{Statistical reporting.}
For the main split-based analyses we report paired split-level means and
two-sided 95\% Student-$t$ intervals. These characterize variation across a small,
non-i.i.d.\ set of official splits, not sampling uncertainty over videos. For the
supplementary standard 50\,Salads long-term-anticipation re-implementation we also
report paired per-video intervals over the 50 pooled test videos, alongside the
five-split intervals because videos within a split share a trained checkpoint. We do not treat the full collection of secondary diagnostic intervals as a familywise hypothesis-testing battery: primary contrasts are specified by the audit design, while secondary intervals are descriptive uncertainty summaries. Implementation
details, full ablations, and all secondary tables are in the supplementary
material.

\section{Results}

\subsection{Primary audit and identification controls}
In the primary four-condition Breakfast audit, video-only, recognizer-history,
GT-history, and oracle context-only score $21.7\%$, $30.3\%$, $62.3\%$, and
$58.8\%$, respectively. Table~\ref{tab:breakfast-main} restores the full
single-step metrics. The near-match between oracle context-only and the full
GT-history condition ($58.8\%$ vs.\ $62.3\%$) is especially informative:
most of the oracle-condition Top-1 score is recoverable without the
observation-window visual input. This does not by itself establish that provenance
is the cause, because procedural context may genuinely be predictive; the matched
recognizer-context experiment below isolates that distinction.

\begin{table}[t]
\centering
\small
\caption{\textbf{Primary four-condition Breakfast audit} (mean$\pm$std over
4 splits). The boundary-aware recognizer defines the primary matched non-oracle
regime; the stricter boundary-free row is shown separately because it changes the
history-recovery pipeline $R$.}
\label{tab:breakfast-main}
\begin{tabular}{lccc}
\toprule
Condition & Act.@1 & Act.@5 & Verb@1 \\
\midrule
Video only
& $21.7{\scriptstyle\pm2.7}$ & $54.3{\scriptstyle\pm2.3}$ & $36.3{\scriptstyle\pm1.3}$ \\
Recognizer history, boundary-free
& $26.3{\scriptstyle\pm0.9}$ & --- & --- \\
Recognizer history, boundary-aware
& $30.3{\scriptstyle\pm3.1}$ & $67.7{\scriptstyle\pm1.5}$ & $44.1{\scriptstyle\pm1.5}$ \\
GT transcript \emph{(oracle)}
& $62.3{\scriptstyle\pm2.6}$ & $89.8{\scriptstyle\pm1.7}$ & $68.1{\scriptstyle\pm2.1}$ \\
Context-only \emph{(oracle context)}
& $58.8{\scriptstyle\pm2.0}$ & $85.4{\scriptstyle\pm1.4}$ & $63.1{\scriptstyle\pm2.8}$ \\
\bottomrule
\end{tabular}
\end{table}

For the boundary-aware recognizer,
$\Delta_{\mathrm{PH}}(R_{\mathrm{BA}})=+32.0{\scriptstyle\pm4.9}$ with
paired 95\% Student-$t$ interval $[+24.1,+39.8]$; using boundary-free history
recognition yields a larger $+36.0$-point gap. These values should not be collapsed
into a single dataset property: they differ because the non-oracle recovery
pipeline differs.

Table~\ref{tab:keycontrols} then contains the four pieces of evidence that carry
the main identification argument. The primary $+32.0$ points compare matched,
separately trained information regimes. The fixed-checkpoint row then shows that
GT history helps even when the recognizer-trained weights are held fixed. The
no-video row shows that high context-only accuracy is not explained merely by
context being predictive: oracle provenance contributes a large additional gap.
The boundary-independent row shows that annotated query/history boundaries are not
a sufficient explanation on Breakfast.

\begin{table*}[t]
\centering
\scriptsize
\setlength{\tabcolsep}{3pt}
\caption{\textbf{Key Breakfast controls.} Each row addresses a different
alternative explanation. The no-video provenance probe is separately retrained, so
its oracle-context score ($59.2$) need not equal the main context-only baseline
($58.8$).}
\label{tab:keycontrols}
\begin{tabular}{p{0.20\textwidth}p{0.12\textwidth}p{0.12\textwidth}p{0.19\textwidth}p{0.29\textwidth}}
\toprule
Comparison & Non-oracle & Oracle & Gap [95\% CI] & Interpretation \\
\midrule
Matched information regimes
& $30.3{\scriptstyle\pm3.1}$
& $62.3{\scriptstyle\pm2.6}$
& $+32.0$ $[+24.1,+39.8]$
& Primary estimator-relative protocol-level audit gap. \\
Fixed checkpoint
& $30.3{\scriptstyle\pm3.1}$
& $44.9{\scriptstyle\pm4.6}$
& $+14.6$ $[+11.0,+18.3]$
& Test-time oracle substitution with weights fixed. \\
No-video context
& $32.5{\scriptstyle\pm1.7}$
& $59.2{\scriptstyle\pm2.5}$
& $+26.8$ $[+22.3,+31.2]$
& Separates oracle provenance from context predictability. \\
Boundary-independent query/history
& $28.1{\scriptstyle\pm1.9}$
& $47.5{\scriptstyle\pm1.8}$
& $+19.4$ $[+16.4,+22.4]$
& Removes annotated boundaries from query scheduling and recovered history. \\
\bottomrule
\end{tabular}
\end{table*}

The full $2\times2$ intervention further reveals a train--test provenance
interaction: the four cell means are $30.3$ (train-rec/test-rec), $44.9$
(train-rec/test-GT), $23.8$ (train-GT/test-rec), and $62.3$
(train-GT/test-GT). Holding the recognizer-trained checkpoint fixed isolates the
clean $+14.6$ test-time substitution contrast. The other row shows why there is no
unique additive ``training contribution'': changing the training regime from
recognizer to GT history raises performance under GT test history
($44.9\rightarrow62.3$), but lowers performance under recognizer test history
($30.3\rightarrow23.8$). Thus, training and test provenance interact strongly.
We therefore use $+32.0$ as the matched protocol-level information-regime gap and
$+14.6$ as the fixed-checkpoint test-time contrast, rather than interpreting the
difference between them as a causal training effect.

\begin{figure}[t]
\centering
\includegraphics[width=0.80\linewidth]{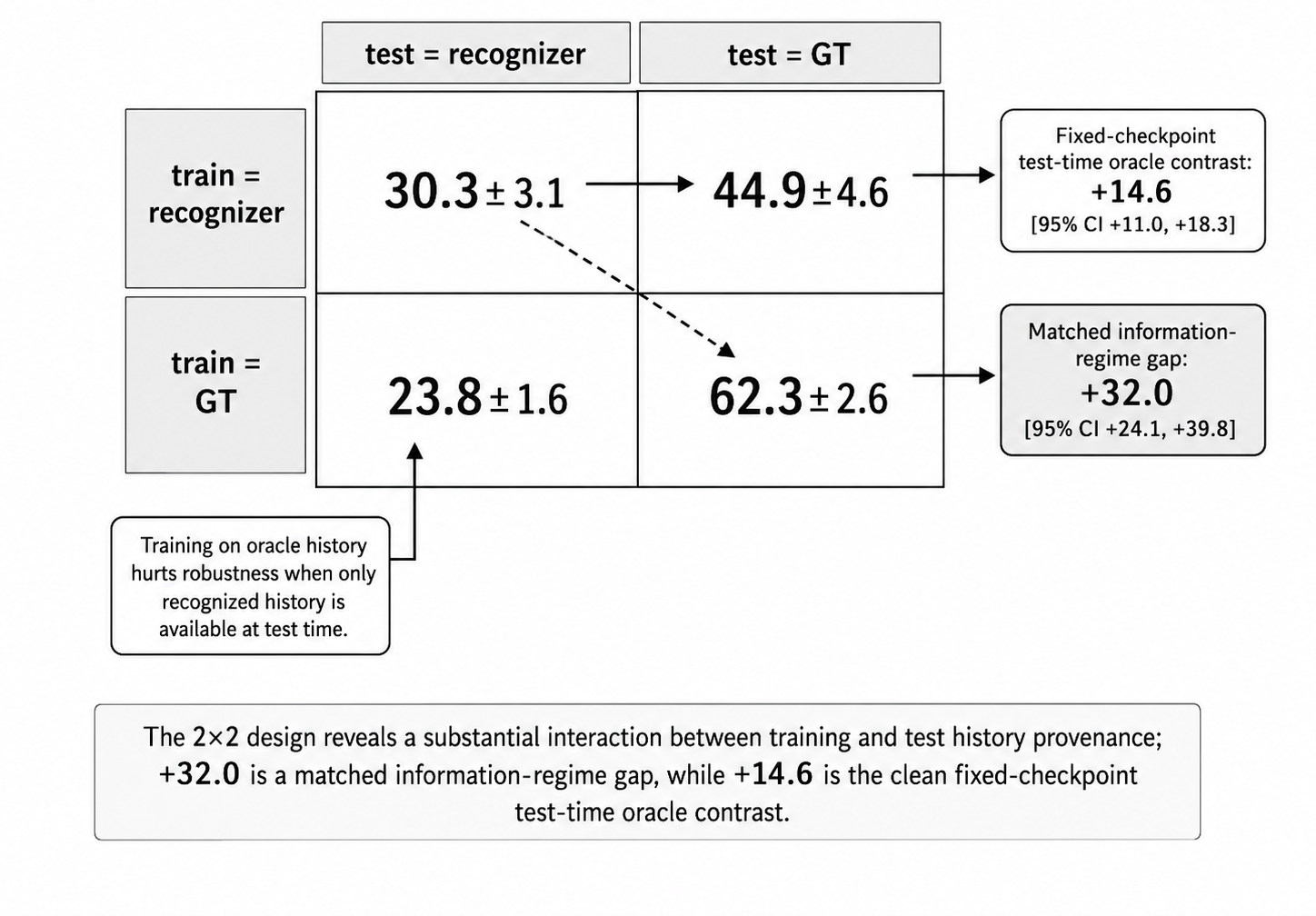}
\caption{\textbf{Train--test provenance intervention.} Holding the
recognizer-trained checkpoint fixed yields a $+14.6$ test-time oracle contrast
($[+11.0,+18.3]$); the off-diagonal cells expose the provenance interaction.}
\label{fig:2x2}
\vspace{-12pt}
\end{figure}

\subsection{Boundary and recovery sensitivity}
The main boundary-aware recognizer uses annotated segment boundaries for the
\emph{observed past}, which itself supplies a mild oracle advantage. To test how
much this choice matters, the boundary-free recognizer removes GT segment
boundaries from history labeling and instead predicts fixed windows. This lowers
the non-oracle anticipation score on all three datasets where the control is run,
thereby widening rather than eliminating the oracle gap
(Table~\ref{tab:bfr-main}). Importantly, this control changes only how observed
history is recovered; the separate fixed-schedule experiment in
Table~\ref{tab:keycontrols} additionally removes annotated boundaries from query
timing on Breakfast.

\begin{table}[t]
\centering
\small
\caption{\textbf{Boundary information in history recovery.} Window-recognition
accuracy measures the past-window labeler itself; BFR and BA are downstream
anticipation Top-1 with boundary-free and boundary-aware recovered histories.}
\label{tab:bfr-main}
\begin{tabular}{lcccc}
\toprule
Dataset & Window-rec. & BFR & BA & GT \\
\midrule
Breakfast & $44.4{\scriptstyle\pm1.2}$ & $26.3{\scriptstyle\pm0.9}$
& $30.3{\scriptstyle\pm3.1}$ & $62.3$ \\
50\,Salads & $74.7{\scriptstyle\pm0.7}$ & $43.2{\scriptstyle\pm4.3}$
& $46.0{\scriptstyle\pm3.5}$ & $53.7$ \\
GTEA & $99.6{\scriptstyle\pm0.1}$ & $66.2{\scriptstyle\pm2.7}$
& $70.2{\scriptstyle\pm4.2}$ & $74.2$ \\
\bottomrule
\end{tabular}
\end{table}

The resulting BFR oracle gaps are $+36.0$, $+10.5$, and $+8.0$ points on
Breakfast, 50\,Salads, and GTEA. The fact that the Breakfast gap remains large
after removing GT segment boundaries from history recovery, and remains
$+19.4$ under the boundary-independent query/history control, makes a pure
boundary-oracle explanation insufficient.

\subsection{Scope and external validity}
The gap is heterogeneous across evaluation settings
(Table~\ref{tab:scope}): it is large on Breakfast, smaller but positive on
50\,Salads and GTEA, compatible with zero on Assembly101, and near zero in the
low-accuracy EPIC-100 diagnostic. The important finding is therefore not that
ground-truth information always produces a large performance boost. Rather,
ground-truth intermediate information can change what an evaluation measures, with
the magnitude depending on the task and recovery process. This motivates explicit
audits of whether a protocol supplies information that the deployed system would
instead need to infer.

\begin{table*}[t]
\centering
\small
\caption{\textbf{Cross-setting scope test.} GTEA, Assembly101, and EPIC-100
($^\dagger$) use diagnostic next-action settings rather than the datasets'
standard anticipation benchmarks. EPIC-100 ($^\ddagger$) is a single-run
low-accuracy diagnostic.}
\label{tab:scope}
\begin{tabular}{lccccl}
\toprule
Setting & Recognizer & GT & $\Delta_{\mathrm{PH}}$ & 95\% split-level interval & Scope \\
\midrule
Breakfast & $30.3{\scriptstyle\pm3.1}$ & $62.3{\scriptstyle\pm2.6}$
& $+32.0$ & $[+24.1,+39.8]$ & primary audit \\
50\,Salads & $46.0{\scriptstyle\pm3.5}$ & $53.7{\scriptstyle\pm2.4}$
& $+7.7$ & $[+5.4,+9.9]$ & next-action audit setting \\
GTEA$^\dagger$ & $70.2{\scriptstyle\pm4.2}$ & $74.2{\scriptstyle\pm3.8}$
& $+4.0$ & $[+1.4,+6.6]$ & diagnostic next-action \\
Assembly101$^\dagger$ & $18.9{\scriptstyle\pm4.4}$ & $19.4{\scriptstyle\pm3.9}$
& $+0.4$ & $[-3.4,+4.2]$ & diagnostic next-action \\
EPIC-100$^{\dagger\ddagger}$ & $14.7$ & $15.0$
& $+0.3$ & --- & diagnostic, $n=1$ \\
\bottomrule
\end{tabular}
\end{table*}

Simple transcript statistics do not explain the ordering: bigram Top-1 is
$49.9/55.5/51.0/55.3\%$ and $H(a_{i+1}\mid a_i)$ is
$1.95/1.55/2.00/1.22$ bits on
Breakfast/50\,Salads/GTEA/Assembly101. Across the three settings with
boundary-free history recognition, higher fixed-window recognition accuracy
($44.4,74.7,99.6\%$) coincides with a smaller BFR gap
($36.0,10.5,8.0$ points), but with only three heterogeneous settings this is
descriptive rather than explanatory.

\paragraph{Standard 50\,Salads protocol.}
We additionally re-implement the published observe-$\alpha$/predict-$\beta$
long-term anticipation structure with $\alpha=0.2$ and
$\beta\in\{0.1,0.2,0.3,0.5\}$. Table~\ref{tab:lta-mini} shows the
GT-minus-recognizer MoC gaps. Per-video intervals exclude zero at 3/4 horizons,
whereas the more conservative five-split intervals include zero throughout.
Because absolute MoC is low and this is not an official checkpoint, we treat it as
supporting \emph{within-re-implementation} evidence rather than a benchmark-level
effect-size estimate. Its role is external-validity corroboration, not replacement
of the primary Breakfast estimand.

\begin{table}[t]
\centering
\scriptsize
\caption{\textbf{Standard 50\,Salads LTA re-implementation.} Point gaps are
GT minus recognizer MoC. The supplement reports the corresponding condition means
and full protocol details.}
\label{tab:lta-mini}
\setlength{\tabcolsep}{3.5pt}
\begin{tabular}{lcccc}
\toprule
 & $\beta=.1$ & $\beta=.2$ & $\beta=.3$ & $\beta=.5$ \\
\midrule
$\Delta_{\mathrm{PH}}$ & $+2.0$ & $+5.2$ & $+3.8$ & $+2.4$ \\
95\% CI (per video) & $[-3.6,+7.5]$ & $[+0.5,+9.8]$ & $[+0.4,+7.1]$ & $[+0.3,+4.4]$ \\
95\% CI (per split) & $[-8.3,+12.3]$ & $[-4.0,+14.3]$ & $[-2.5,+10.0]$ & $[-1.5,+6.2]$ \\
\bottomrule
\end{tabular}
\end{table}

\subsection{Sensitivity and mechanism}
The audit is strongly estimator-relative. In a controlled recognizer-quality sweep
on Breakfast, recognizer-history Top-1 rises from $26.4\%$ to $31.5\%$ as
injected history noise decreases, so a better recovery pipeline can shrink the
measured gap. A separate history-dose experiment shows the converse leverage of
oracle context: within that setup, adding a single GT past-action label raises
accuracy from $23.8\%$ to $51.0\%$, whereas one recovered label reaches
$27.4\%$.

\begin{figure*}[t]
\centering
\includegraphics[width=0.77\textwidth]{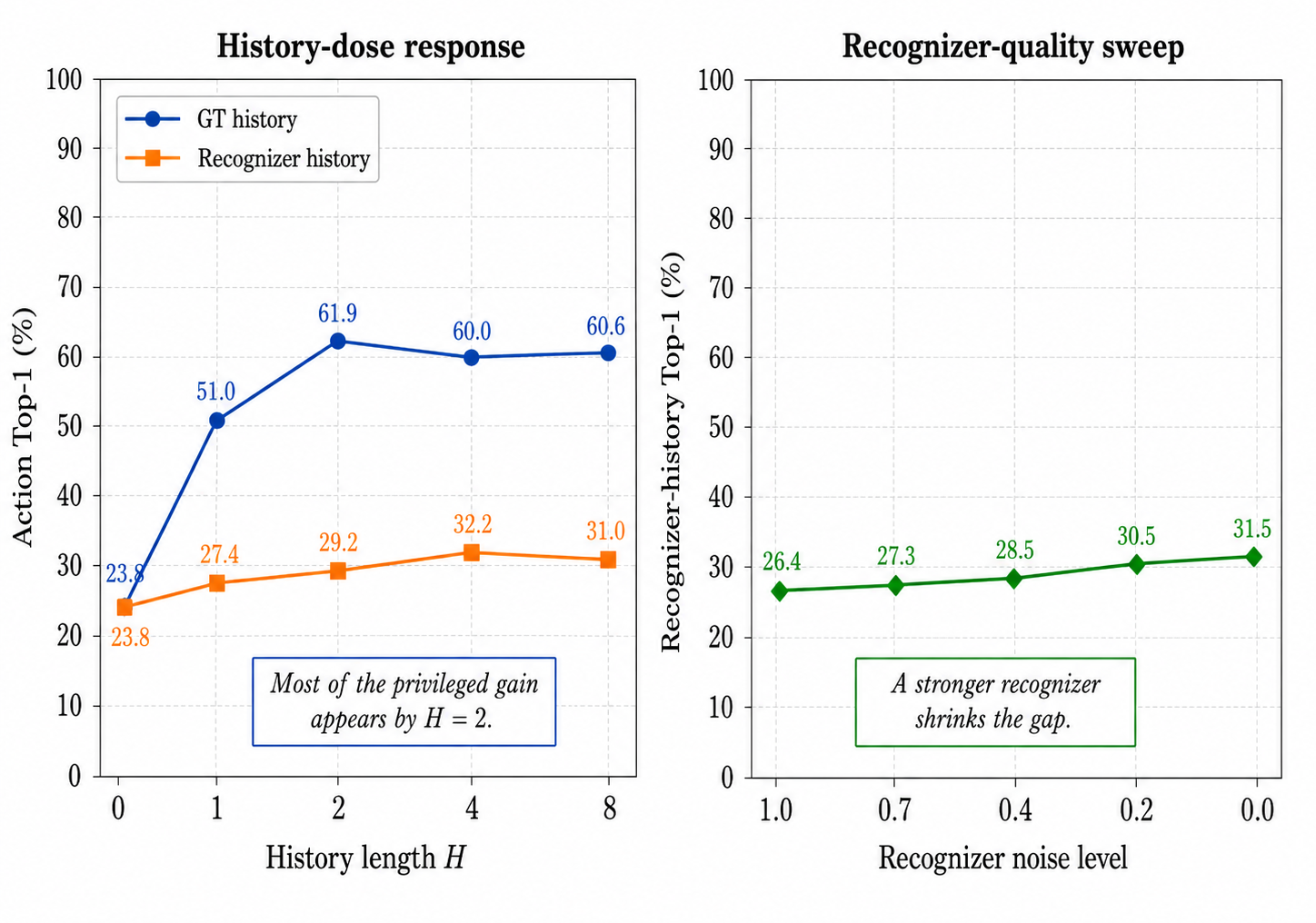}
\caption{\textbf{Estimator-relative sensitivity on Breakfast.}
\emph{Left:} history-dose response. \emph{Right:} recognizer-history performance
as controlled history noise decreases. The second panel directly demonstrates
that improving recovery changes the matched non-oracle score and therefore the
measured $\Delta_{\mathrm{PH}}(R)$.}
\label{fig:sensitivity}
\end{figure*}

The effect is not confined to one-step prediction. On a dedicated Breakfast
multi-horizon model, $\Delta_{\mathrm{PH}}$ is
$+25.9,+17.1,+13.7,+11.5,+12.1$ for horizons $1$--$5$, with all paired
split-level intervals above zero. This is a separately trained horizon-specific
experiment, so its $h=1$ values need not reproduce the single-step model exactly.
We treat the pattern as horizon-wise persistence rather than a causal explanation
of the decay; full condition rows and intervals are reported in
Supplementary Table~\ref{tab:sup-dense}.

\begin{figure}[!t]
\centering
\includegraphics[width=0.77\linewidth]{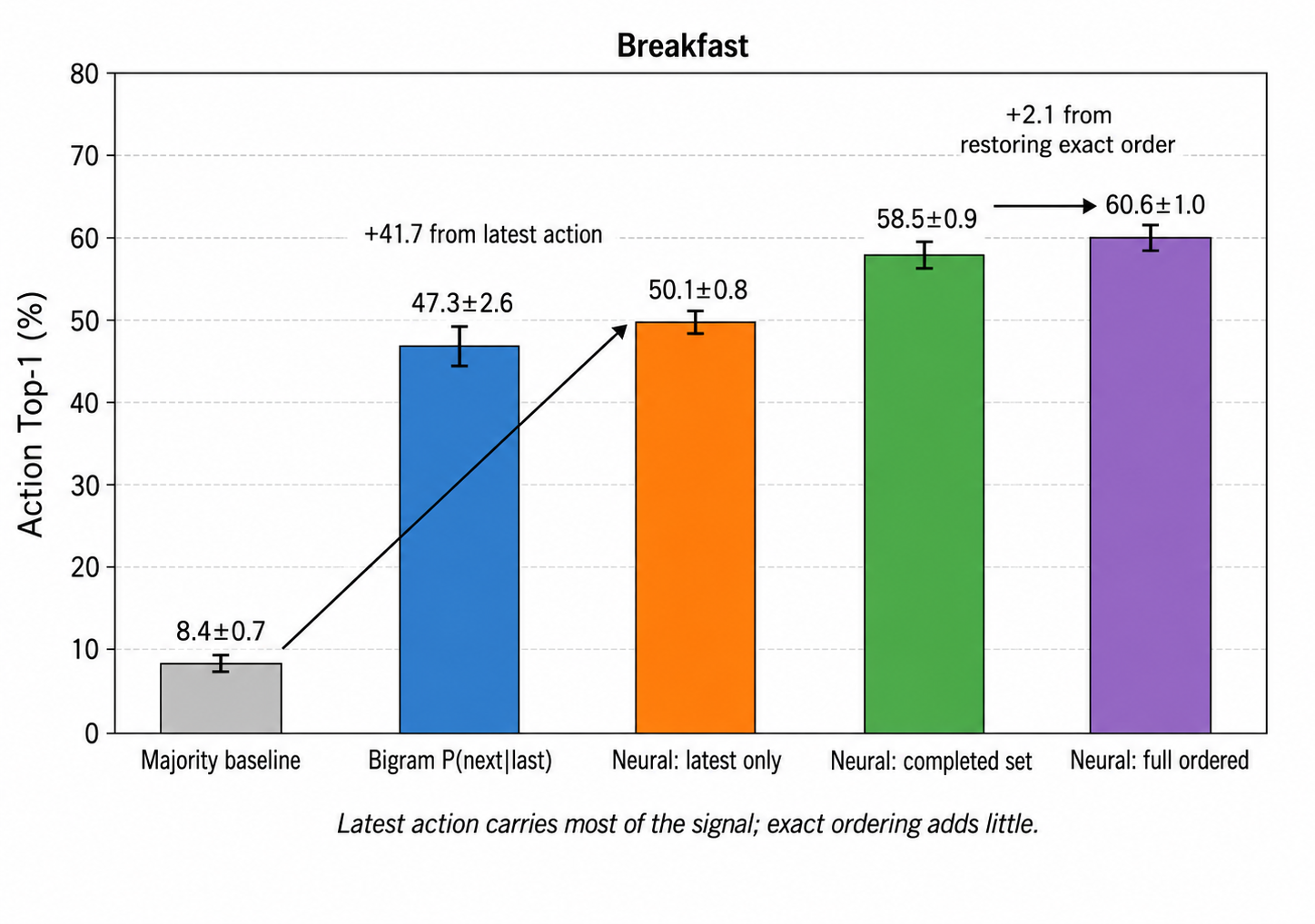}
\caption{\textbf{Breakfast decomposition.} The latest action alone captures a
large fraction of the context-only gain; exact ordering adds comparatively little.}
\label{fig:decomp}
\end{figure}

The decomposition in Fig.~\ref{fig:decomp} localizes much of the privileged signal.
On Breakfast, a majority baseline is $8.4\%$, a count-based bigram using the latest
action reaches $47.3\%$, a neural latest-action predictor reaches $50.1\%$, the
completed-action set reaches $58.5\%$, and the full ordered history reaches
$60.6\%$. Thus, local action context carries much of the signal, while restoring
exact order adds only $+2.1$ points on Breakfast. In separately retrained
context-only ablations, symbolic state adds only $+0.1$ once action history is
present. Full three-dataset decomposition, state ablations, multi-horizon results,
error analysis, attribution, and architecture checks are supplementary.

\section{Discussion and Recommendations}
The central lesson is evaluation-theoretic rather than specific to action history:
a benchmark may remain temporally causal and target-leakage-free while evaluating
a different capability because it supplies an oracle intermediate representation.
The evidence above is intentionally layered: no single number carries the entire
claim.

\begin{table}[H]
\centering
\scriptsize
\caption{\textbf{Evidence hierarchy.} Each control targets a different rival
explanation.}
\label{tab:hierarchy}
\begin{tabular}{p{0.31\linewidth}p{0.59\linewidth}}
\toprule
Evidence & What it addresses \\
\midrule
Fixed checkpoint: $+14.6$
& Not only a consequence of training under cleaner history. \\
Matched regimes: $+32.0$
& Primary estimator-relative audit gap. \\
No-video provenance: $+26.8$
& Not merely ``context is predictive.'' \\
Boundary-independent: $+19.4$
& Annotated query/history boundaries are not sufficient. \\
50\,Salads standard LTA
& Supporting evidence beyond audit-specific next-action construction. \\
\bottomrule
\end{tabular}
\end{table}

The practical implication is reporting discipline, not removal of symbolic
history. Oracle-conditioned scores can be useful upper bounds, ablations, or
capability probes, but they should not be interpreted as measuring the same
construct as a system that must recover the intermediate representation from
observations. Conversely, privilege is deployment-relative rather than intrinsic to
a representation: if the intended deployment genuinely provides an authoritative
action log, then a GT-like history is not privileged for that deployment and the
relevant non-oracle comparison changes accordingly.

This distinction is also separate from classical temporal leakage. A protocol may
contain no future frames and no target label, yet still provide an oracle
representation of the observed past. Temporal validity asks whether information
from after prediction time enters the model; the present audit asks whether the
representation of information available \emph{before} prediction time is itself
oracle. Both checks are necessary for interpreting what a benchmark score
supports.

Finally, the measured quantity is deliberately estimator-relative. A weak recovery
pipeline can make the oracle/non-oracle gap look large; a stronger one can shrink
it. Our controlled noise sweep and the BFR comparison demonstrate that dependence
empirically. The audit therefore requires reporting the recovery procedure $R$
alongside $\Delta_{\mathrm{PH}}(R)$ rather than treating the gap as a permanent
property of a dataset.

\paragraph{Reusable checklist.}
Before interpreting a score as evidence of the intended capability, ask:
(1) are all inputs available without oracle annotation;
(2) are they temporally valid;
(3) are intermediate variables supplied from ground truth;
(4) what recovery pipeline $R$ would produce them non-oracularly;
(5) how well does context alone predict the target;
(6) how sensitive is the gap to recovery quality or boundary information; and
(7) are matched non-oracle and oracle regimes reported separately?

\paragraph{Limitations.}
External validity is strengthened by the standard-protocol 50\,Salads
re-implementation, but broader replication under unchanged standard protocols
remains open; GTEA, Assembly101, and EPIC-100 are still diagnostic settings.
Accordingly, the paper's general contribution is the audit framework and the
within-setting identification logic, not a claim that anticipation benchmarks share
a universal effect size.
The recovery-quality sweep is controlled rather than a benchmark of the strongest
modern recognizers, so stronger recovery could shrink
$\Delta_{\mathrm{PH}}(R)$ further. Cross-setting comparisons confound backbone,
label space, visual content, recovery difficulty, and procedure structure, and we
do not infer a causal explanation for heterogeneity. Multi-horizon and most
mechanistic controls remain Breakfast-centered. For 50\,Salads, per-video
intervals exclude zero at 3/4 horizons while split-level intervals include zero
throughout; because videos within a split share a checkpoint, the split-level
summary is the more conservative characterization.

\FloatBarrier
\bibliographystyle{plainnat}
\bibliography{references}

% ============================================================
% Supplementary material / appendix begins AFTER the references.
% ============================================================
\clearpage
\appendix
\section*{Supplementary Material}
% Compact appendix float spacing: keep related tables visually grouped.
\setlength{\textfloatsep}{7pt plus 2pt minus 2pt}
\setlength{\floatsep}{7pt plus 2pt minus 2pt}
\setlength{\intextsep}{7pt plus 2pt minus 2pt}
\addcontentsline{toc}{section}{Supplementary Material}

\section{Detailed Related Work and Protocol Context}
Prior work spans clean/GT observations, recognized histories, direct visual
inference, and language-derived symbolic histories. Table~\ref{tab:prior_protocols}
summarizes examples that motivate treating provenance as an explicit evaluation
variable rather than a fixed modeling choice. We do not claim that prior work
necessarily misinterpreted oracle-conditioned scores; the audit asks which
capability interpretation a score supports when the oracle variable is unavailable
under the intended deployment setting.

\begin{table}[h]
\centering
\small
\caption{\textbf{Examples of observed-past representations in prior work.}}
\label{tab:prior_protocols}
\begin{tabular}{p{0.25\linewidth}p{0.25\linewidth}p{0.40\linewidth}}
\toprule
Work & Observed-past source & Relevance \\
\midrule
Uncertainty-aware anticipation~\cite{farha2019uncertainty}
& Clean/GT and recognized observations
& Separates clean and inferred observed actions. \\
Cycle-consistent anticipation~\cite{farha2020cycle}
& Visual features
& Motivates direct video-to-future prediction rather than fixed transcripts. \\
AntGPT~\cite{zhao2024antgpt}
& Symbolic actions derived from observations
& Uses semantic action sequences as context. \\
AAG~\cite{benavent2025aag}
& GT, recognizer-, or VLM-derived history
& Makes history provenance a first-order modeling choice. \\
\bottomrule
\end{tabular}
\end{table}

\begin{figure}[h]
\centering
\begin{tikzpicture}
\node[inner sep=0] (taskfig) {\includegraphics[width=0.95\linewidth]{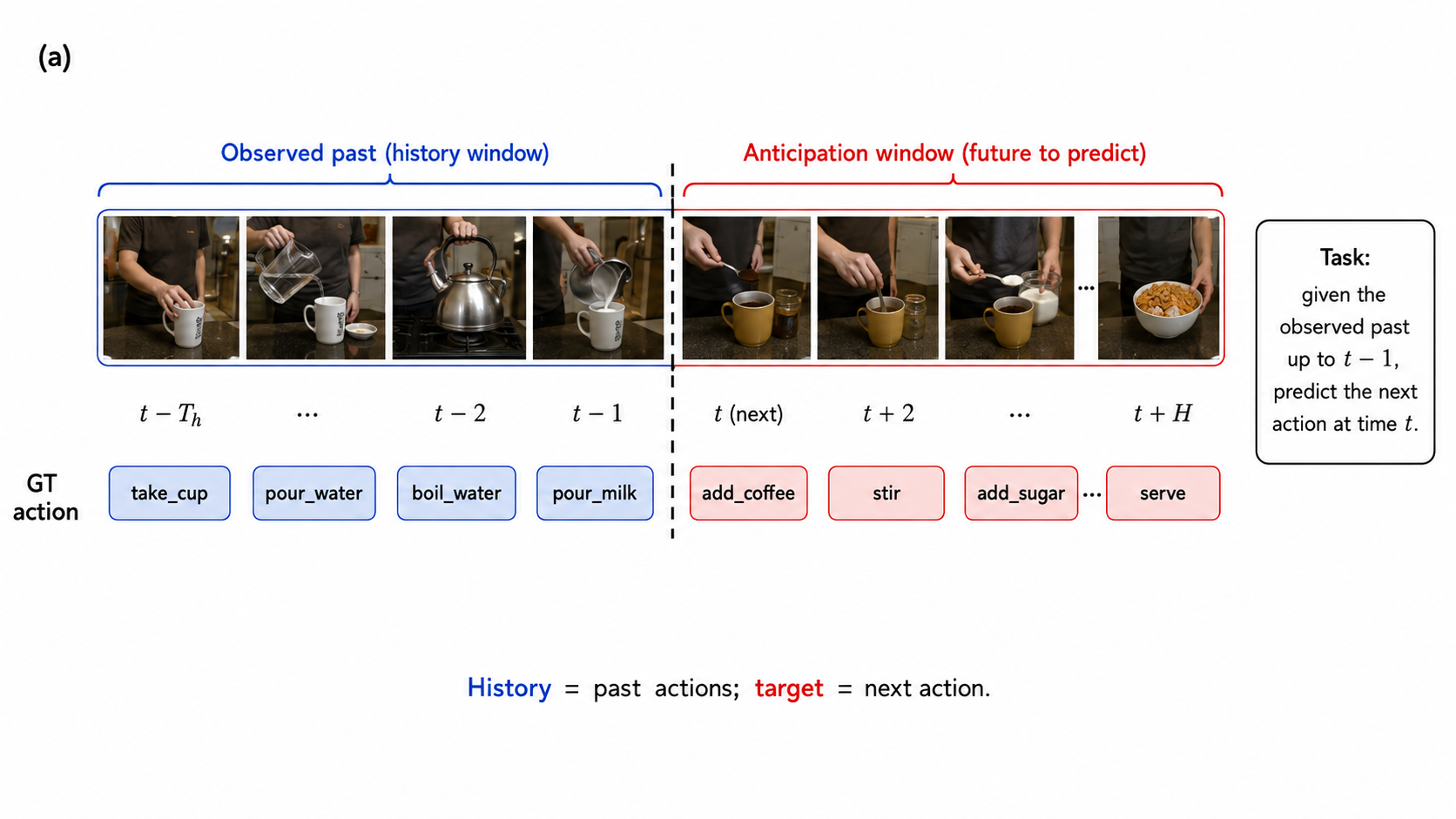}};
% Erase the obsolete "(a)" marker and any small raster remnants around it.
\fill[white] ([xshift=-0.08cm,yshift=0.08cm]taskfig.north west)
             rectangle ([xshift=1.45cm,yshift=-1.10cm]taskfig.north west);
\end{tikzpicture}
\caption{\textbf{Task definition.} The model observes only the procedural prefix
available before prediction time. Future frames and the target label are never
model inputs.}
\label{fig:task}
\end{figure}

\begin{figure}[h]
\centering
\includegraphics[width=0.82\linewidth]{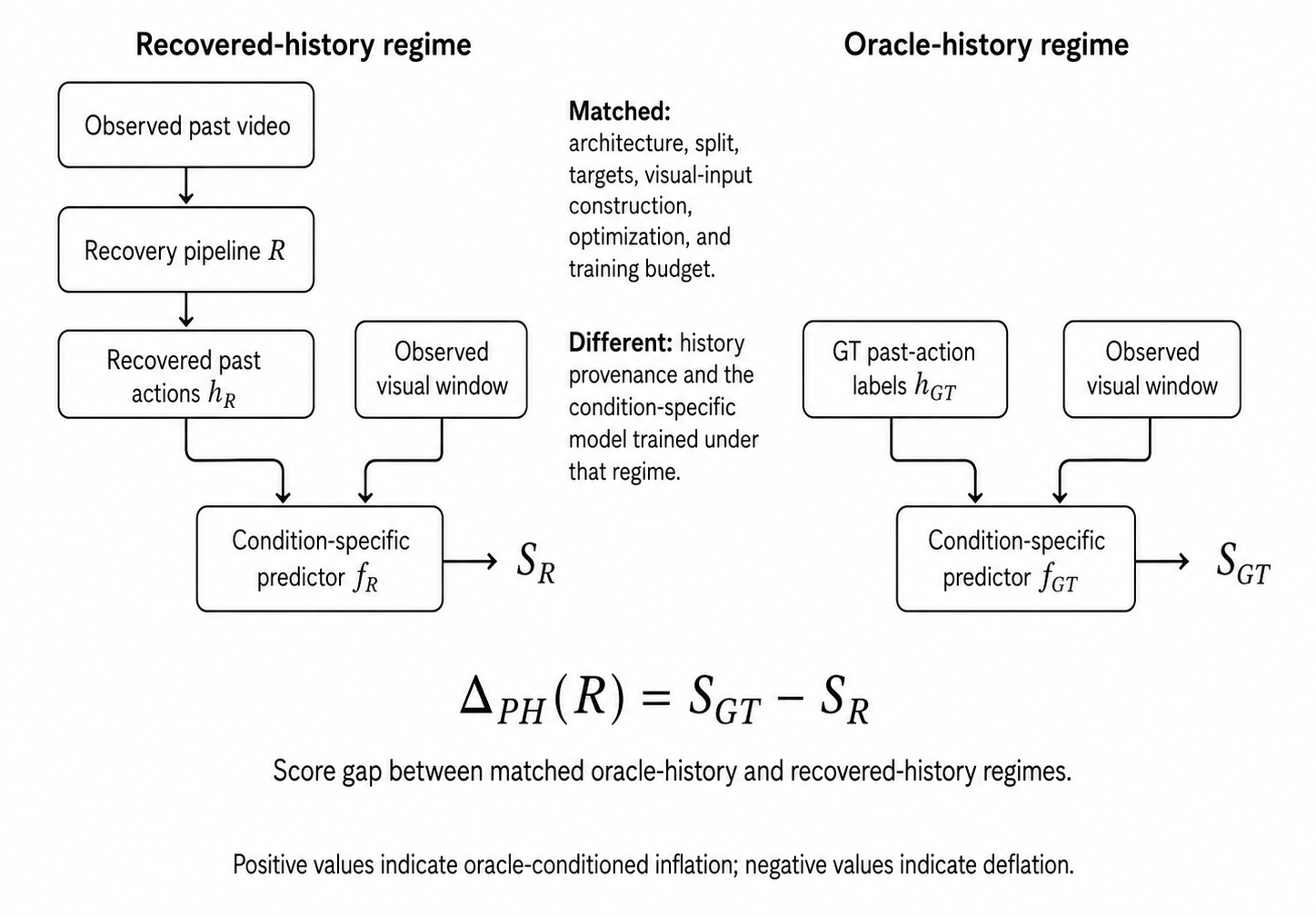}
\caption{\textbf{Operationalization of the estimator-relative privileged-information gap.} The matched
oracle-history and recovered-history regimes share task, architecture family,
targets, visual construction, optimization, and training budget, while history
provenance differs.}
\label{fig:pii}
\end{figure}

\section{Implementation and Reproducibility Details}
Throughout the supplement, \emph{non-oracle} denotes information recovered from
ordinary system inputs under the stated protocol; \emph{oracle} denotes
annotation-supplied information unavailable under that deployment claim. The
reference anticipator projects 2048-d I3D/TSM features to $d=256$, encodes the
observation window with a 2-layer transformer encoder (4 heads, GELU, dropout
0.1), and mean-pools valid frames. Past actions use $|\mathcal{A}|+2$ embeddings
(PAD/SIL included) and a GRU. The symbolic state is a deterministic binary summary
of held-object, per-object processed/contained, and appliance-state indicators,
updated only from observed past actions. Visual, history, and state representations
are concatenated, passed through linear+LayerNorm+GELU fusion, and read out by
action, verb, and noun heads.

Training uses AdamW, weight decay $10^{-4}$, learning rate $3\times10^{-4}$,
batch size 128, and 8 epochs. Action cross-entropy is combined with verb and noun
auxiliary losses weighted by 0.3. The segment recognizer is a 3-layer MLP over
mean-pooled segment features. Boundary-free history recognition labels 16-feature-frame
windows with stride 8 feature frames and collapses consecutive duplicates. The anticipation
observation window contains the last 32 feature frames before the prediction boundary.
GT-history and recognizer-history models use the same architecture, optimization,
splits, targets, visual construction, and training budget but are trained under
their corresponding history regimes.

For the fixed-schedule control, queries are issued on the extracted feature
sequence starting at feature-frame 64 after a 64-feature-frame warm-up and then
every 60 feature frames. The target is the first action starting after the query,
and recognizer history is produced by the boundary-free fixed-window recognizer. No annotated
segment boundary is used as a model input, to schedule the query, or to construct
recognizer history; annotations are used only to define evaluation targets.
For the fixed-schedule run, the recovered symbolic history is truncated/padded to
the eight most recent action labels ($H=8$).

\paragraph{Statistical conventions.}
Primary audit contrasts use paired Student-$t$ intervals over official splits.
These intervals summarize split-to-split variation rather than sampling
uncertainty over videos. For the standard 50\,Salads LTA re-implementation we
also report paired per-video intervals over the 50 pooled test videos, alongside
the five-split intervals because videos within a split share a trained checkpoint.
We do not treat the collection of secondary diagnostic intervals as a familywise
hypothesis-testing battery; secondary intervals are descriptive uncertainty
summaries unless explicitly identified as a primary audit contrast.

\section{Full Breakfast Results}
This section expands the compact key-control presentation in the main paper.
Table~\ref{tab:sup-main} restores the full single-step metrics, while
Tables~\ref{tab:sup-2x2}--\ref{tab:sup-query} provide the full values underlying
the fixed-checkpoint and boundary-independent controls.

\begin{table}[h]\centering\small
\caption{\textbf{Breakfast next-action anticipation} (4 splits).}
\label{tab:sup-main}
\begin{tabular}{lccc}
\toprule
Condition & Act.@1 & Act.@5 & Verb@1 \\
\midrule
Video only & $21.7{\scriptstyle\pm2.7}$ & $54.3{\scriptstyle\pm2.3}$ & $36.3{\scriptstyle\pm1.3}$ \\
Recognizer history, boundary-free & $26.3{\scriptstyle\pm0.9}$ & --- & --- \\
Recognizer history, boundary-aware & $30.3{\scriptstyle\pm3.1}$ & $67.7{\scriptstyle\pm1.5}$ & $44.1{\scriptstyle\pm1.5}$ \\
GT transcript (oracle) & $62.3{\scriptstyle\pm2.6}$ & $89.8{\scriptstyle\pm1.7}$ & $68.1{\scriptstyle\pm2.1}$ \\
Context-only oracle probe & $58.8{\scriptstyle\pm2.0}$ & $85.4{\scriptstyle\pm1.4}$ & $63.1{\scriptstyle\pm2.8}$ \\
\bottomrule
\end{tabular}
\end{table}

\begin{table}[h]\centering\small
\caption{\textbf{$2\times2$ train-history$\times$test-history intervention}
(Breakfast, 4 splits).}
\label{tab:sup-2x2}
\begin{tabular}{lcc}
\toprule
Train history & Test: recognizer & Test: GT \\
\midrule
Recognizer & $30.3{\scriptstyle\pm3.1}$ & $44.9{\scriptstyle\pm4.6}$ \\
GT oracle & $23.8{\scriptstyle\pm1.6}$ & $62.3{\scriptstyle\pm2.6}$ \\
\bottomrule
\end{tabular}
\end{table}

\section{Boundary and Recovery Sensitivity}
This section expands the main-paper sensitivity discussion. The boundary table
separates annotation use in history recovery from the fixed-schedule query control,
while the dose and recovery-quality results quantify the estimator-relative nature
of $\Delta_{\mathrm{PH}}(R)$.

The first control below removes annotated boundaries from both query scheduling
and recovered-history construction; the next two tables isolate boundary use in
history recognition and the effect of history length.

\begin{table}[H]\centering\small
\caption{\textbf{Boundary-independent query and history-recovery control}
(Breakfast, 4 splits). Queries start at feature-frame 64 and repeat every 60
feature frames; the visual window is 32 feature frames, and the boundary-free
history recognizer uses 16-feature-frame windows with stride 8.}
\label{tab:sup-query}
\begin{tabular}{lc}
\toprule
Condition & Act.@1 (\%) \\
\midrule
Video only & $21.5{\scriptstyle\pm1.2}$ \\
Recognizer history & $28.1{\scriptstyle\pm1.9}$ \\
GT history (oracle) & $47.5{\scriptstyle\pm1.8}$ \\
\midrule
$\Delta_{\mathrm{PH}}$ & $+19.4{\scriptstyle\pm1.9}$ \\
95\% Student-$t$ interval & $[+16.4,+22.4]$ \\
\bottomrule
\end{tabular}
\end{table}

\begin{table}[H]\centering\small
\caption{\textbf{Boundary-free versus boundary-aware history recognition.}}
\label{tab:sup-bfr}
\begin{tabular}{lccccc}
\toprule
Dataset & Window-recog. & BFR & BA & Drop & GT \\
\midrule
Breakfast & $44.4{\scriptstyle\pm1.2}$ & $26.3{\scriptstyle\pm0.9}$
& $30.3{\scriptstyle\pm3.1}$ & $-4.0$ & $62.3$ \\
50\,Salads & $74.7{\scriptstyle\pm0.7}$ & $43.2{\scriptstyle\pm4.3}$
& $46.0{\scriptstyle\pm3.5}$ & $-2.8$ & $53.7$ \\
GTEA & $99.6{\scriptstyle\pm0.1}$ & $66.2{\scriptstyle\pm2.7}$
& $70.2{\scriptstyle\pm4.2}$ & $-4.0$ & $74.2$ \\
\bottomrule
\end{tabular}
\end{table}

Boundary-free history recognition lowers the recovered-history anticipation score
by $4.0$, $2.8$, and $4.0$ points on Breakfast, 50\,Salads, and GTEA,
respectively, but does not remove the oracle gap. The following dose experiment
then varies only how much symbolic history is available on Breakfast.

\begin{table}[H]\centering\small
\caption{\textbf{History-length dose response on Breakfast.}}
\label{tab:sup-dose}
\begin{tabular}{lccccc}
\toprule
History length $H$ & 0 & 1 & 2 & 4 & 8 \\
\midrule
GT history & 23.8 & 51.0 & 61.9 & 60.0 & 60.6 \\
Recognizer history & 23.8 & 27.4 & 29.2 & 32.2 & 31.0 \\
\bottomrule
\end{tabular}
\end{table}

\paragraph{Recognizer-quality sweep.}
For controlled noise levels $1.0,0.7,0.4,0.2,0.0$, recognizer-history anticipation
Top-1 is $26.4,27.3,28.5,30.5,31.5\%$, respectively.

\FloatBarrier
\section{Cross-Setting Details}
This section provides the full columns omitted from the main-paper scope table.
The 50\,Salads row below is the boundary-aligned next-action audit setting; it
should not be read as the published long-term-anticipation benchmark. The separate
standard long-term-anticipation re-implementation is reported in
Sec.~\ref{sec:sup-lta}.

\begin{table}[h]\centering\scriptsize
\caption{\textbf{Full cross-setting audit.} The 50\,Salads row is the
next-action audit setting; the standard long-term-anticipation re-implementation
is reported separately below.}
\label{tab:sup-cross}
\begin{tabular}{lcccccccc}
\toprule
Dataset & Backbone & \#cls & Video & Rec.(BA) & Rec.(BFR) & GT & $\Delta^{BA}_{PH}$ & $\Delta^{BFR}_{PH}$ \\
\midrule
Breakfast & I3D & 47 & 21.7 & $30.3{\scriptstyle\pm3.1}$ & $26.3{\scriptstyle\pm0.9}$ & $62.3{\scriptstyle\pm2.6}$ & $+32.0$ & $+36.0$ \\
50\,Salads & I3D & 19 & 40.7 & $46.0{\scriptstyle\pm3.5}$ & $43.2{\scriptstyle\pm4.3}$ & $53.7{\scriptstyle\pm2.4}$ & $+7.7$ & $+10.5$ \\
GTEA$^\dagger$ & I3D & 11 & 67.3 & $70.2{\scriptstyle\pm4.2}$ & $66.2{\scriptstyle\pm2.7}$ & $74.2{\scriptstyle\pm3.8}$ & $+4.0$ & $+8.0$ \\
Assembly101$^\dagger$ & TSM & 86 & 21.3 & $18.9{\scriptstyle\pm4.4}$ & --- & $19.4{\scriptstyle\pm3.9}$ & $+0.4$ & --- \\
EPIC-100$^{\dagger\ddagger}$ & TSN & 331 & 14.4 & 14.7 & --- & 15.0 & $+0.3$ & --- \\
\bottomrule
\end{tabular}
\end{table}

Paired split-level BA intervals are:
Breakfast $[+24.1,+39.8]$;
50\,Salads $[+5.4,+9.9]$;
GTEA $[+1.4,+6.6]$;
Assembly101 $[-3.4,+4.2]$.
EPIC-100 is a single run. GTEA, Assembly101, and EPIC-100 are diagnostic
next-action settings, not official standard anticipation benchmarks.

\subsection{Standard long-term anticipation on 50\,Salads}
\label{sec:sup-lta}
This expands the abbreviated external-validity result in the main paper. It is
supporting evidence beyond the audit-specific next-action construction, not a
replacement for the primary Breakfast estimand. Point gaps are positive at every
tested prediction ratio; per-video intervals exclude zero at three horizons,
whereas the more conservative split-level intervals include zero throughout.

\begin{table}[h]\centering\small
\caption{\textbf{50\,Salads under the standard long-term anticipation protocol}
(observe $20\%$, predict $\beta$; MoC). Gaps are computed from unrounded paired
values and may differ slightly from subtracting displayed rounded means.}
\label{tab:sup-lta}
\begin{tabular}{lcccc}
\toprule
Observed history & $\beta=.1$ & $\beta=.2$ & $\beta=.3$ & $\beta=.5$ \\
\midrule
Recognizer & 7.6 & 6.2 & 4.9 & 4.1 \\
GT transcript & 9.5 & 11.3 & 8.7 & 6.5 \\
\midrule
$\Delta_{\mathrm{PH}}$ & +2.0 & +5.2 & +3.8 & +2.4 \\
95\% CI (per video) & $[-3.6,+7.5]$ & $[+0.5,+9.8]$ & $[+0.4,+7.1]$ & $[+0.3,+4.4]$ \\
95\% CI (per split) & $[-8.3,+12.3]$ & $[-4.0,+14.3]$ & $[-2.5,+10.0]$ & $[-1.5,+6.2]$ \\
\bottomrule
\end{tabular}
\end{table}

The per-video analysis pools 50 test-video paired differences across the five
splits. Because videos within a split share a trained checkpoint, the five-split
analysis is the more conservative summary of between-training-split variation.

\FloatBarrier
\section{Multi-Horizon and Decomposition}
These results expand the main paper's short horizon-persistence and mechanism
summaries. The multi-horizon table reports all paired intervals, and the
three-dataset decomposition shows where the latest-action, completed-set, and
ordered-history representations agree or differ.

\begin{table}[h]\centering\small
\caption{\textbf{Breakfast multi-horizon anticipation} (4 splits).}
\label{tab:sup-dense}
\begin{tabular}{lccccc}
\toprule
Condition & $h=1$ & $h=2$ & $h=3$ & $h=4$ & $h=5$ \\
\midrule
Video only & 25.4 & 23.9 & 21.7 & 19.8 & 20.6 \\
Recognizer history & 35.1 & 29.5 & 24.4 & 21.8 & 22.0 \\
GT history & 61.0 & 46.6 & 38.1 & 33.3 & 34.0 \\
\midrule
$\Delta_{\mathrm{PH}}$ & +25.9 & +17.1 & +13.7 & +11.5 & +12.1 \\
95\% interval & $[+22.8,+29.1]$ & $[+14.9,+19.3]$ & $[+11.1,+16.3]$ &
$[+9.9,+13.0]$ & $[+10.3,+13.9]$ \\
\bottomrule
\end{tabular}
\end{table}

\begin{table}[h]\centering\small
\caption{\textbf{Decomposition of privileged-history signal} (GT history, no
observation-window video).}
\label{tab:sup-decomp}
\begin{tabular}{lccc}
\toprule
Predictor & Breakfast & 50\,Salads & GTEA \\
\midrule
Majority & $8.4{\scriptstyle\pm0.7}$ & $7.7{\scriptstyle\pm0.8}$ & $26.6{\scriptstyle\pm0.4}$ \\
Bigram $P(a_{i+1}|a_i)$ & $47.3{\scriptstyle\pm2.6}$ & $47.5{\scriptstyle\pm1.8}$ & $50.6{\scriptstyle\pm3.9}$ \\
Latest action only & $50.1{\scriptstyle\pm0.8}$ & $60.7{\scriptstyle\pm3.6}$ & $73.8{\scriptstyle\pm6.0}$ \\
Completed-action set & $58.5{\scriptstyle\pm0.9}$ & $54.5{\scriptstyle\pm3.8}$ & $73.6{\scriptstyle\pm3.6}$ \\
Full ordered history & $60.6{\scriptstyle\pm1.0}$ & $54.5{\scriptstyle\pm3.3}$ & $73.3{\scriptstyle\pm4.5}$ \\
\midrule
Gain from latest action & +41.7 & +53.0 & +47.2 \\
Gain from exact order & +2.1 & $\approx0$ & -0.4 \\
\bottomrule
\end{tabular}
\end{table}

Paired contrasts: on 50\,Salads, completed-set minus latest-only is
$-6.2{\scriptstyle\pm4.2}$ with $[-11.4,-1.0]$, while order minus set is
approximately zero with $[-1.9,+1.9]$. On GTEA, set minus latest is
$-0.2{\scriptstyle\pm4.0}$ with $[-6.6,+6.2]$ and order minus set is
$-0.4{\scriptstyle\pm1.3}$ with $[-2.4,+1.7]$. The Breakfast shuffle control is
$60.6{\scriptstyle\pm1.0}$ ordered versus $58.5{\scriptstyle\pm0.9}$ shuffled.

\FloatBarrier
\section{Context-Only Ablations}
This section expands the no-video provenance result summarized in the main
paper. Table~\ref{tab:sup-ablation} isolates the source of context-only
performance: history alone nearly matches history+state, so the auxiliary symbolic
state is not responsible for the main probe result. Table~\ref{tab:sup-recprobe}
then holds the absence of observation-window video fixed and replaces recovered
context with oracle context, directly testing provenance. The two experiments are
separately retrained and are therefore interpreted within their matched setups.

\begin{table}[h]\centering\small
\caption{\textbf{Source of context-only probe signal on Breakfast.}}
\label{tab:sup-ablation}
\begin{tabular}{lc}
\toprule
Probe input & Act.@1 (\%) \\
\midrule
History only & $59.1{\scriptstyle\pm1.8}$ \\
State only & $46.5{\scriptstyle\pm2.1}$ \\
History + state & $59.2{\scriptstyle\pm2.2}$ \\
\bottomrule
\end{tabular}
\end{table}

\begin{table}[h]\centering\small
\caption{\textbf{No-video provenance probe on Breakfast} (4 splits). This is a
separately retrained experiment; its GT score need not equal the main $58.8\%$
context-only baseline.}
\label{tab:sup-recprobe}
\begin{tabular}{lc}
\toprule
Probe context & Act.@1 (\%) \\
\midrule
Recognizer context & $32.5{\scriptstyle\pm1.7}$ \\
GT context & $59.2{\scriptstyle\pm2.5}$ \\
\midrule
Privileged-context gap & $+26.8{\scriptstyle\pm2.8}$ \\
95\% Student-$t$ interval & $[+22.3,+31.2]$ \\
\bottomrule
\end{tabular}
\end{table}

\FloatBarrier
\section{Architecture Sensitivity}
Table~\ref{tab:sup-arch} reports the full Breakfast-only architecture sensitivity
check summarized in the main paper. The reference, TCN, and GRU rows use four
splits; SSM, causal-LM, and diffusion use two splits and are treated only as
secondary robustness checks rather than architecture-independent evidence. The
purpose is to reduce concern about a reference-model artifact, not to rank model
families or claim architecture-independent robustness.

\begin{table}[h]\centering\small
\caption{\textbf{Breakfast-only architecture sensitivity.}}
\label{tab:sup-arch}
\begin{tabular}{lcccccc}
\toprule
Architecture & Video & Rec. & GT & $\Delta_{\mathrm{PH}}$ & $n$ & CI/status \\
\midrule
Transformer-fusion & 21.7 & 30.3 & 62.3 & +32.0 & 4 & $[+24.1,+39.8]$ \\
TCN & 24.9 & 32.2 & 59.3 & +27.1 & 4 & $[+24.8,+29.4]$ \\
GRU & 27.8 & 37.0 & 61.7 & +24.8 & 4 & $[+20.2,+29.4]$ \\
SSM & 24.4 & 29.4 & 57.6 & +28.2 & 2 & robustness \\
Causal LM & 21.6 & 32.1 & 57.6 & +25.6 & 2 & robustness \\
Diffusion & 18.4 & 28.8 & 57.8 & +28.9 & 2 & robustness \\
\bottomrule
\end{tabular}
\end{table}

\FloatBarrier
\section{Error Analysis and Attribution}
Tables~\ref{tab:sup-errors} and~\ref{tab:sup-attribution} are diagnostics on
Breakfast split~1 that expand the mechanism discussion in the main paper. They
localize transition-conditioned performance, recognizer-error propagation, and
input-stream reliance, but are not used as population-level causal evidence or as
a decomposition of the full oracle gap.

\begin{table}[h]\centering\small
\caption{\textbf{Error analysis under privileged history} (Breakfast split 1).}
\label{tab:sup-errors}
\begin{tabular}{lccc}
\toprule
\multicolumn{4}{l}{Target-class frequency} \\
\midrule
Bin & rarest & middle & commonest \\
Mean train count & 102.6 & 192.6 & 411.8 \\
Act.@1 & 50.2 & 62.5 & 77.5 \\
\midrule
\multicolumn{4}{l}{Transition probability $P(a_{i+1}|a_i)$} \\
\midrule
Bin & rare & medium & frequent \\
Mean probability & 0.028 & 0.306 & 0.782 \\
Act.@1 & 19.1 & 76.8 & 94.4 \\
\midrule
\multicolumn{4}{l}{Recognizer-error propagation} \\
\midrule
Latest action & correct & incorrect & \\
$n$ & 447 & 793 & \\
Act.@1 & 51.5 & 20.6 & \\
\bottomrule
\end{tabular}
\end{table}

\begin{table}[h]\centering\small
\caption{\textbf{Input-stream attribution} (Breakfast split 1).}
\label{tab:sup-attribution}
\begin{tabular}{lcc}
\toprule
 & GT history & Recognizer history \\
\midrule
Full model & 63.4 & 31.7 \\
Zero video & $39.0(-24.4)$ & $12.1(-19.6)$ \\
Zero history & $14.2(-49.2)$ & $9.3(-22.4)$ \\
Zero state & $61.6(-1.8)$ & $29.5(-2.2)$ \\
\midrule
grad$\times$input video & 57.3\% & 79.5\% \\
grad$\times$input history & 36.0\% & 17.8\% \\
grad$\times$input state & 6.7\% & 2.7\% \\
\bottomrule
\end{tabular}
\end{table}

\FloatBarrier
\section{Additional EPIC Diagnostic}
The main paper uses the 331-class EPIC-KITCHENS-100 diagnostic
(video/recognizer/GT $=14.4/14.7/15.0$, $\Delta_{\mathrm{PH}}=+0.3$).
A supplementary 104-class variant gives recognizer accuracy $63.4\%$,
video-only Top-1 $19.4\%$, recognizer-history $19.5\%$, GT-history $19.7\%$,
and $\Delta_{\mathrm{PH}}=+0.2$.

\FloatBarrier
\end{document}